\documentclass[letterpaper, 10 pt, conference]{ieeeconf}  

\IEEEoverridecommandlockouts                              

\usepackage{graphics} 
\usepackage{epsfig} 
\usepackage{mathptmx} 
\usepackage{times} 
\usepackage{amsmath} 
\usepackage{amssymb}  
\usepackage{amsfonts}
\usepackage{mathrsfs} 

\usepackage{threeparttable}
\usepackage{textcomp}
\usepackage{siunitx}
\usepackage{multirow}
\usepackage{adjustbox}
\usepackage{diagbox}
\usepackage{booktabs, multirow}
\usepackage{adjustbox}
\usepackage{booktabs, multirow}
\usepackage{xcolor}
\usepackage[utf8]{inputenc}
\usepackage{graphicx}
\usepackage{tikz}
\usepackage{listings}                       
\usepackage{algorithm, algorithmic}                      
\usetikzlibrary{backgrounds}

\usepackage[noadjust]{cite}

\usepackage{gensymb}

\title{\LARGE \bf CDIS : Cross-Dimensional Class-Agnostic 3D Instance Segmentation\\ via 2D Mask Tracking and 3D-2D Projection Merging
}
\author{Juno Kim$^{1*}$ Hye-Jung Yoon$^{1*}$ Yesol Park$^{1*} $ Byoung-Tak Zhang$^{1}$
\thanks{This work was partly supported by the IITP (RS-2021-II212068-AIHub/10\%, RS-2021-II211343-GSAI/15\%, RS-2022-II220951-LBA/15\%, RS-2022-II220953-PICA/20\%), NRF (RS-2024-00353991-SPARC/20\%, RS-2023-00274280-HEI/10\%), and KEIT (RS-2024-00423940/10\%) grant funded by the Korean government.}
\thanks{*Authors have equal contributions}
\thanks{$^{1}$Interdisciplinary Program in AI, Seoul National University}%
}

\begin{document}
\maketitle
\thispagestyle{empty}
\pagestyle{empty}


\begin{abstract}
Class-agnostic 3D instance segmentation is critical for robotic systems operating in unknown environments, enabling perception of previously unseen objects for reliable manipulation and navigation. Existing approaches typically project per-frame 2D instance masks into 3D and merge them, which often breaks object identities across time and yields fragmented 3D instances. We introduce Cross-Dimensional Class-Agnostic 3D Instance Segmentation (CDIS), a zero-shot framework that explicitly tracks 2D instance masks across frames and associates them with 3D superpoints, creating a feedback loop between 2D and 3D. This cross-dimensional reasoning links temporally stable 2D tracks with spatially coherent 3D regions, producing globally consistent 3D instance labels without any 3D-specific training. Experiments on benchmark datasets demonstrate that CDIS achieves higher accuracy and consistency than state-of-the-art zero-shot methods, while remaining efficient and scalable to diverse real-world environments.
\end{abstract}

\section{Introduction}
Understanding the 3D environment is essential for robotic manipulation and navigation in real-world scenarios. Class-agnostic 3D instance segmentation is particularly crucial when robots encounter previously unseen objects, enabling robust manipulation, path planning, and obstacle avoidance without requiring prior knowledge of object categories.

Traditional approaches to 3D instance segmentation rely heavily on large, annotated 3D datasets~\cite{rozenberszki2022language, yeshwanth2023scannet++, chang2017matterport3d, straub2019replica}. This dependency severely limits generalization to novel objects and reduces applicability in real-world robotic environments~\cite{schult2023mask3d, takmaz2023openmask3d, kolodiazhnyi2024oneformer3d}. To overcome these limitations, recent research has shifted toward leveraging 2D instance segmentation models within 3D perception pipelines~\cite{lu2023ovir, rozenberszki2024unscene3d, xu2023sampro3d, yan2024maskclustering, yang2023sam3d, guo2023sam, kim2024ov, yin2024sai3d, nguyen2024open3dis, huang2023segment3d, huang2024openins3d}. These projection-based methods capitalize on the generalization capabilities of 2D models trained on large-scale annotated datasets due to advancements in large 2D segmentation models~\cite{cheng2021per, cheng2022masked, kirillov2023segment, qi2022high, park2025dafusion}. For instance, models like SAM3D~\cite{yang2023sam3d} project 2D segmentation masks from the Segment Anything Model (SAM)~\cite{kirillov2023segment} and CropFormer~\cite{qi2022high} into 3D space, then iteratively merge the projected masks based on geometric overlap.

However, projection-based approaches still face significant challenges in real-world robotic scenarios. Depth data often suffer from noise, missing regions, or calibration imperfections, especially for reflective or distant surfaces, thereby amplifying segmentation errors in individual frames. When these frame-level errors are directly projected into 3D and merged, they frequently break object identities across time and produce fragmented 3D instances. Although some methods attempt to refine depth estimates using reconstructed point clouds~\cite{guo2023sam}, these pipelines still struggle to maintain consistent instance labels across frames.

\begin{figure} [t!]
\begin{center}
\includegraphics[width=0.9\columnwidth]{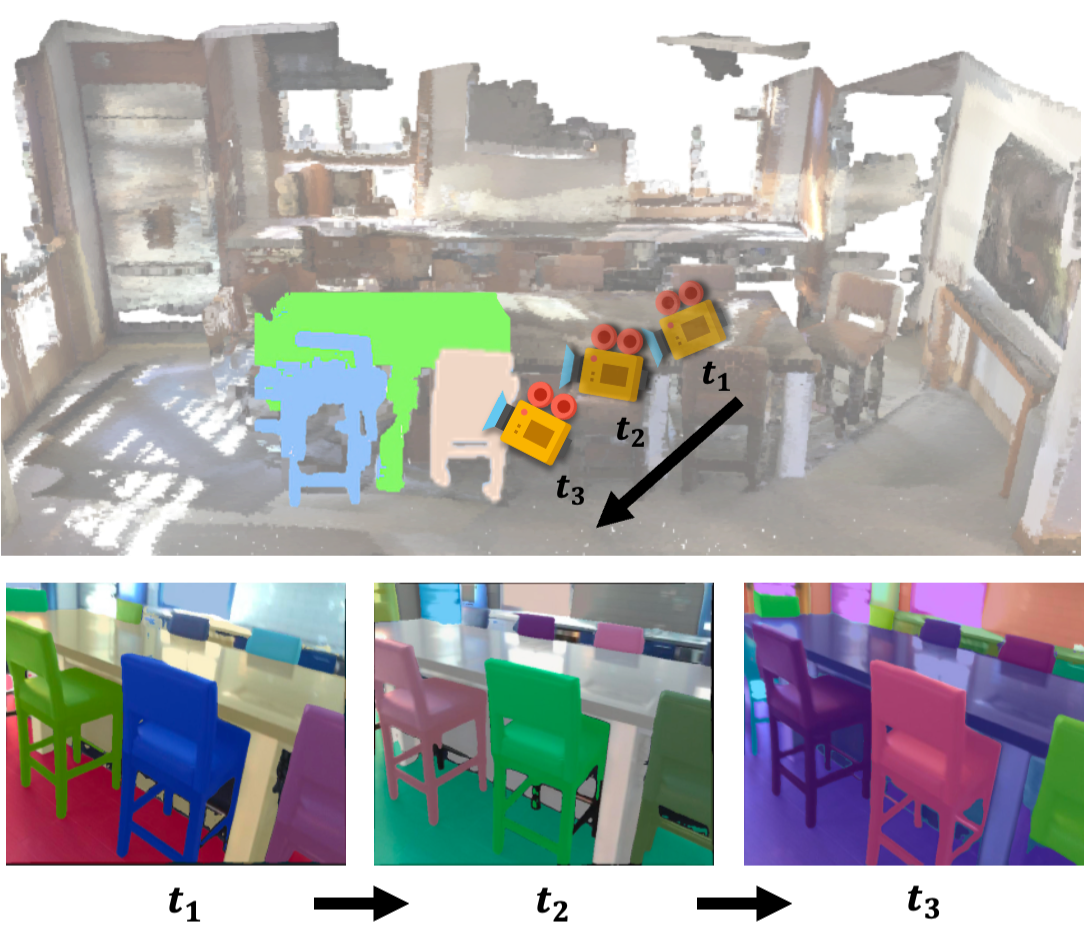}

\caption{\textbf{Overview of CDIS.} Top image illustrates a reconstructed real-world 3D scene generated from sequentially captured RGB-D frames at timestamps \( t_1, t_2, t_3 \). The bottom images show the progression of 2D instance segmentation results over time, demonstrating our method, which improves segmentation quality across frames through mask tracking. CDIS integrates segmentation cues from both 2D and 3D spaces to enhance class-agnostic 3D instance segmentation and improve alignment across multiple views.}
\label{fig_system}
\vspace{-7mm}
\end{center}
\end{figure}

\begin{figure*} [t!]
\begin{center}
\vspace{+2mm}
\includegraphics[width=\textwidth]{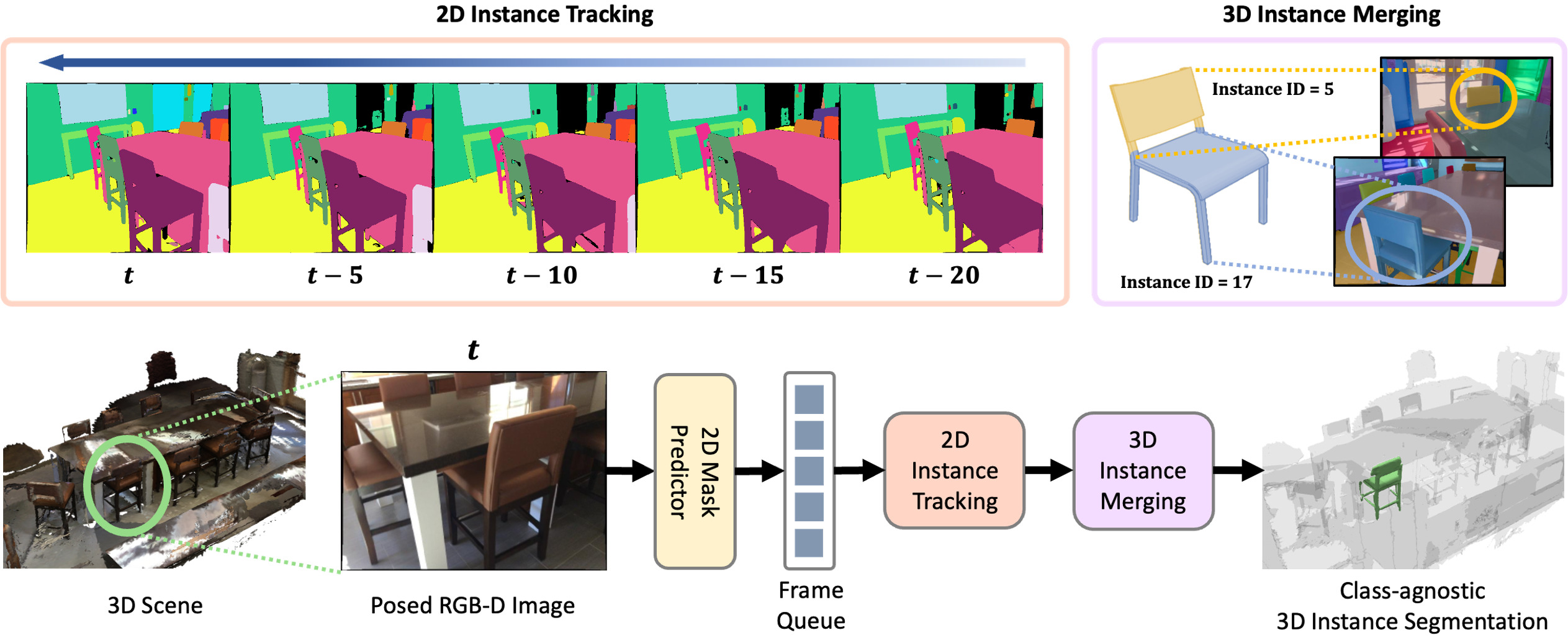}
\vspace{-2mm}
\caption{\textbf{Overall Framework of CDIS.} 
The proposed pipeline for zero-shot, class-agnostic 3D instance segmentation. 
Posed RGB-D frames are input into a 2D mask predictor to generate instance masks for each frame, 
which are tracked over time using depth-based projection, frame warping, and 2D IoU matching. 
Tracked masks are then associated with pre-computed 3D superpoints, enabling spatio-temporal merging 
of instances across frames based on geometric consistency. 
Finally, duplicate superpoint assignments are resolved through overlap and temporal co-occurrence analysis, 
resulting in a unified and consistent 3D instance segmentation.}
\label{fig_overview}
\vspace{-5mm}
\end{center}
\end{figure*}

In this paper, we present Cross-Dimensional Class-Agnostic 3D Instance Segmentation (CDIS), a zero-shot framework designed to overcome the limitations of projection-based pipelines by jointly reasoning across 2D and 3D domains, as illustrated in Fig.~\ref{fig_system}. CDIS first establishes temporally consistent 2D instance tracks from sequential RGB frames via geometric warping and multi-frame matching. This reduces short-term segmentation errors that often arise when processing frames independently. These 2D tracks are then aligned with precomputed 3D superpoints, which serve as geometrically coherent anchors to consolidate fragmented predictions. Through iterative refinement that alternates between 2D appearance cues and 3D structural constraints, CDIS corrects transient segmentation failures, prevents drift and produces globally consistent 3D instance labels without any 3D-specific training.

We refer to this bidirectional reasoning paradigm as \emph{cross-dimensional} processing. Unlike previous approaches that commit early to a fixed 3D representation obtained from single-frame projections, CDIS maintains a feedback loop between 2D and 3D: temporal cues from 2D tracking correct short-term errors, while 3D superpoint structure resolves spatial ambiguities and prevents long-term drift. This design mitigates segmentation fragmentation and improves the robustness of zero-shot 3D instance segmentation.

We evaluate CDIS on publicly available benchmark datasets~\cite{rozenberszki2022language, yeshwanth2023scannet++}, demonstrating superior accuracy and consistency compared to existing zero-shot methods. The results highlight CDIS as a scalable solution for autonomous robotic systems that require reliable perception of novel objects in diverse real-world environments.

\section{Related Works}
\subsection{Learning-Based 3D Instance Segmentation}
Traditional 3D instance segmentation approaches rely heavily on large-scale annotated datasets~\cite{rozenberszki2022language, yeshwanth2023scannet++}. 
Fully supervised models~\cite{schult2023mask3d, kolodiazhnyi2024oneformer3d, hou20193d, ngo2023isbnet} typically adopt 3D convolutional neural networks or transformer-based backbones to process point clouds or voxelized inputs, producing detailed per-instance predictions. 
Although these techniques achieve strong performance in controlled, closed-set environments, they require labor-intensive 3D annotations and often generalize poorly to previously unseen object categories, limiting their use in dynamic or open-world settings.

\subsection{2D-Driven 3D Segmentation}
To reduce this dependency on 3D labels, several methods incorporate pretrained 2D segmentation models into 3D pipelines. 
These 2D-driven approaches~\cite{yang2023sam3d, yin2024sai3d, yan2024maskclustering} project per-frame 2D instance masks into 3D using depth maps and merge them based on geometric overlap or clustering. 
Such pipelines effectively leverage strong 2D visual priors but often remain tied to specific category sets or predefined taxonomies, limiting their scalability to open-world scenarios.

Building on these ideas, class-agnostic and training-free methods have emerged to segment all objects without relying on semantic labels or 3D-specific training. 
For example, OVIR-3D~\cite{lu2023ovir}, SAMPro3D~\cite{xu2023sampro3d}, and Open3DIS~\cite{nguyen2024open3dis} densely segment each RGB-D frame using general-purpose 2D models such as SAM~\cite{kirillov2023segment} or CropFormer~\cite{qi2022high}, then lift the resulting masks into 3D and merge them using superpoint associations or clustering strategies. 
These training-free, class-agnostic pipelines remove the need for labeled 3D data but still process frames independently, making them susceptible to segmentation drift and fragmented object identities over time.

Our CDIS framework belongs to this family of training-free, class-agnostic approaches but addresses the temporal inconsistency issue by explicitly tracking 2D instance masks across frames and using 3D superpoints as geometric anchors, creating a feedback loop between 2D and 3D that yields globally consistent 3D instance labels.

\section{Method} \label{sec}
We present a zero-shot, class-agnostic 3D instance segmentation framework 
that leverages a 2D entity-level segmentation model~\cite{qi2022high}, 
pre-trained on large-scale datasets. 
Given an RGB-D scene represented by a sequence of \(T\) posed frames 
\(\{I_t, D_t, C_t\}_{t=1}^T\), where \(t\) denotes the frame index, 
\(I_t \in \mathbb{R}^{H \times W \times 3}\) is the RGB image, 
\(D_t \in \mathbb{R}^{H \times W}\) is the depth map, 
and \(C_t \in \mathbb{R}^{4 \times 4}\) is the camera pose, 
our objective is to generate 3D instance masks that accurately delineate 
distinct objects within the scene.

\textbf{Overview.}
Our approach generates 3D instance segmentation in three stages 
(Fig.~\ref{fig_overview}). 
First, we perform 2D instance tracking to obtain temporally 
consistent 2D instance labels across frames by projecting 2D instance masks into 3D space using the corresponding depth maps, warping past frames, and matching instances using 2D mask IoU.
Second, we perform 3D-guided 2D instance merging, where 
projected 3D superpoints are used to associate 2D instances across 
all frames based on their geometric consistency, producing spatio-temporally 
consistent 2D instance IDs. 
Finally, we perform 3D instance consolidation to resolve duplicate 
assignments of superpoints to multiple instance IDs by analyzing 
3D overlaps and temporal co-occurrence, producing a unique 
and unified mapping of superpoints to instance IDs. 
The result is a fully spatio-temporally consistent 3D instance segmentation 
without requiring any task-specific training.

\subsection{2D Instance Tracking}

For each frame \(t\), a 2D instance mask \(M_t\) is generated using the 2D segmentation model. 
These masks are then projected into 3D space using the depth map \(D_t\) and camera intrinsic matrix \(K\), producing a set of 3D points 
\(P_t \in \mathbb{R}^{N_t \times 3}\) with corresponding instance IDs 
\(G_t \in \mathbb{Z}^{N_t}\), where \(N_t\) is the number of valid points in frame \(t\).
To ensure temporal continuity, we maintain a frame queue 
\(\mathscr{Q}_t = \{ t_p \mid t - q_{\text{max}} < t_p \leq t \}\), 
storing up to \(q_{\text{max}}\) previous frames for comparison.

\textbf{Frame Warping.} To track instances, we warp the 3D points from a past frame \( t_p \in \mathscr{Q}_t \) into the current frame \( t \) using the relative transformation:

\begin{equation}
C_{t_p \to t} = C_t^{-1} C_{t_p}, \quad
P_{t_p \to t} = C_{t_p \to t} P_{t_p}
\end{equation}

where \( C_t \) and \( C_{t_p} \) are the respective pose matrices representing the camera transformations for frames \( t \) and \( t_p \), and \( P_{t_p} \) represents the homogeneous 3D points from frame \( t_p \). The transformation \( C_{t_p \to t} \) maps these points into the coordinate system of the current frame.
The warped 3D points are then projected back into 2D space:

\begin{equation}
p_{t_p \to t}^{\text{2D}} = \frac{K P_{t_p \to t}^{\top}}{(K P_{t_p \to t}^{\top})_z}
\end{equation}

$p_{t_p \to t}^{\text{2D}}$ denotes the normalized 2D projections, with \((\cdot)_z\) indicating the Z-coordinate (depth) in camera space.

A new mask $M_{t_p \to t}$ is initialized and updated based on valid 2D coordinates 
within the image boundaries:
\begin{equation}
M_{t_p \to t}(x_i, y_i) = G_{t_p}[i] 
\quad \text{if} \quad (x_i, y_i) = p_{t_p \to t}^{\text{2D}}[i],
\end{equation}
with $(x_i, y_i)$ representing the projected pixel coordinate of the $i$-th point 
and $G_{t_p}[i]$ indicating its associated instance label.

\begin{figure} [t!]
\begin{center}
\vspace{+3mm}
\includegraphics[width=\columnwidth]{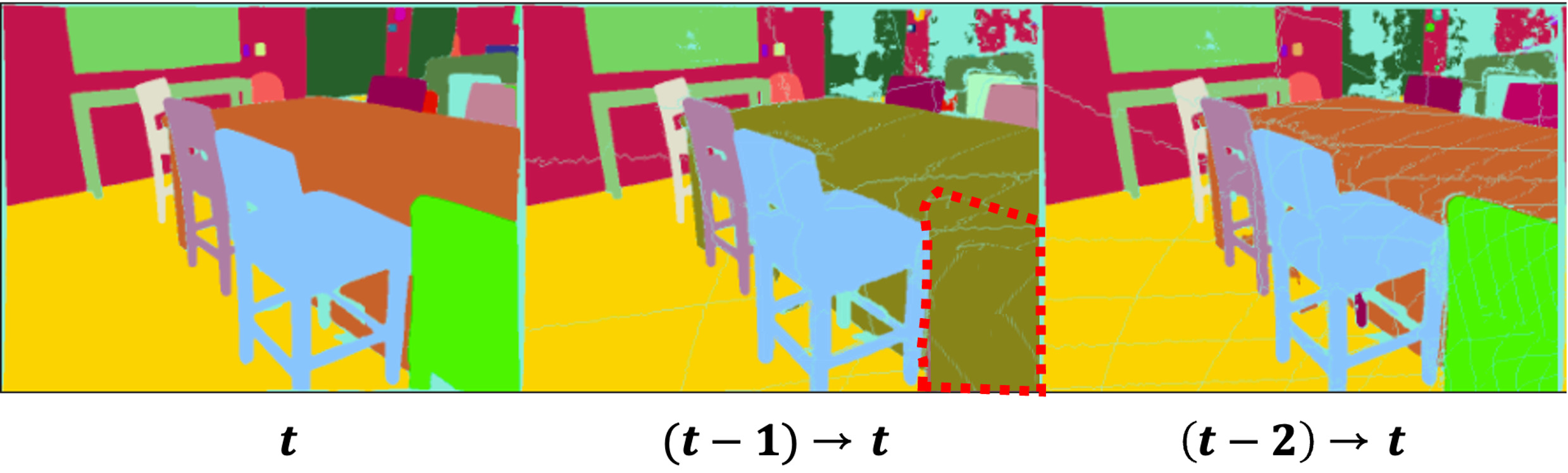} 
\vspace{-3mm}
\caption{\textbf{Handling 2D Segmentation Failures with Multi-Frame Tracking.}
When generating the mask for frame $t$, using only the warped mask from frame $t\!-\!1$ propagates an error where the chair and table are merged (red box). 
CDIS prevents this by also referencing frame $t\!-\!2$, where the objects are correctly separated, and restores proper instance IDs in frame $t$. 
By leveraging multiple past frames rather than only the adjacent frame, CDIS mitigates temporary segmentation errors and prevents drift in object identities.}
\label{fig_2D_grouping}
\vspace{-6mm}
\end{center}
\end{figure}

\textbf{Mask Matching.}
To maintain consistent instance labels across frames, 
we compare the current mask $M_t$ with the warped previous mask $M_{t_p \to t}$ 
at the instance level. 
For each instance label $u$ in $M_t$ and each instance label $v$ in $M_{t_p \to t}$, 
we compute their intersection-over-union (IoU):
\begin{equation}
\text{IoU}(u, v) = 
\frac{|\{M_t = u\} \cap \{M_{t_p \to t} = v\}|}
{|\{M_t = u\} \cup \{M_{t_p \to t} = v\}|},
\end{equation}
where $\{M_t = u\}$ denotes the set of pixels in mask $M_t$ labeled $u$. If any label $v$ exceeds the threshold $\tau_{\text{IoU2D}}$, we set the label of instance $u$ to $v^* = \arg\max_v \text{IoU}(u,v)$.
If no label satisfies the threshold, a new unique instance label is assigned. 
This remapping is applied to all pixels of instance $u$, ensuring temporal consistency across frames (see Fig.~\ref{fig_2D_grouping}). 
After this process, each frame $M_t$ contains instance labels that are 
temporally matched to previous frames, resulting in a sequence 
$\{M_t\}_{t=1}^{T}$ with consistent instance identities across time.

\subsection{3D-Guided 2D Instance Merging}
We next leverage 3D geometry to improve instance consistency. 
A set of pre-computed 3D superpoints is used, where each 
superpoint \(S_s = \{p_i \mid p_i \in \mathbb{R}^3\}\) is a cluster of 3D points 
obtained from~\cite{felzenszwalb2004efficient}. 

\textbf{3D Superpoint Projection.}
Each superpoint is projected onto all frames, producing its corresponding 
2D footprint, denoted as $\Pi_t(s)$. Each superpoint is then associated with the instance label 
in \(M_t\) that maximizes overlap:
\begin{equation}
A_t(s) = 
\arg \max_{u} 
\frac{|\Pi_t(s) \cap (M_t = u)|}{|\Pi_t(s)|}.
\end{equation}

\textbf{Instance Association and Merging.}
For instance labels $u$ (frame $t$) and $v$ (frame $t+1$), 
their similarity is computed directly from their associated superpoints:
\begin{equation}
\text{IoU}_{3D}(u,v) =
\frac{|\{s \mid A_t(s) = u\} \cap \{s \mid A_{t+1}(s) = v\}|}
     {|\{s \mid A_t(s) = u\} \cup \{s \mid A_{t+1}(s) = v\}|}.
\end{equation}
If $\text{IoU}_{3D}(u,v) > \tau_{\text{IoU3D}}$, 
the labels $u$ and $v$ are merged and assigned the same identity. 
This merging is applied hierarchically, pairwise across neighboring frames, 
reducing the number of frame groups by half per iteration, 
until a single consistent set of instance identities is obtained 
for the entire sequence. 
As a result, each frame $M_t$ contains 2D instance IDs 
that are tracked across time and merged based on geometrical information, 
producing spatially and temporally consistent instance labels.

\subsection{3D Instance Consolidation}

Following the previous stage, we obtain spatio-temporally matched instances, 
where each 2D instance ID is associated with one or more 3D superpoints. 
However, duplicate assignments may occur, as multiple instance IDs can share 
the same superpoints. 
To establish a unique and consistent instance ID for each superpoint, 
we perform a consolidation process.

\textbf{Detection of Overlapping Instances.} 
For each pair of instances $(u,v)$, we define their associated superpoint sets as 
$\mathcal{S}_u$ and $\mathcal{S}_v$, where 
$\mathcal{S}_u = \{ s \mid A_t(s) = u \}$ 
denotes the set of superpoints assigned to instance $u$. 
We then compute the 3D intersection-over-minimum (IoMin):
\begin{equation}
\text{IoMin}_{3D}(u,v) = 
\frac{|\mathcal{S}_u \cap \mathcal{S}_v|}
     {\min(|\mathcal{S}_u|,|\mathcal{S}_v|)}.
\end{equation}
Pairs satisfying $\text{IoMin}_{3D}(u,v) > \tau_{\text{IoMin3D}}$ 
are considered overlapping and evaluated further.

\textbf{Temporal Co-Occurrence Analysis.}
To distinguish between different objects that are spatially adjacent 
and identical objects observed at different times, 
we examine the temporal co-occurrence of the two instances:
\begin{equation}
CO(u,v) = 
\frac{|T_u \cap T_v|}{\min(|T_u|,|T_v|)},
\end{equation}
where $T_u$ and $T_v$ are the sets of frames in which instances $u$ and $v$ 
are observed. 
If $CO(u,v) > \tau_{\text{co}}$, the two instances are likely distinct objects 
whose overlap originates from segmentation noise, 
and the overlapping superpoints are removed from the larger instance. 
Otherwise, the two instances are considered views of the same object and are merged.

\textbf{Iterative Refinement and Final Assignment.}
This process is repeated until no instance pair exceeds 
the $\tau_{\text{IoMin3D}}$ threshold. 
Finally, each superpoint is assigned the instance ID 
with which it has been associated most frequently across all frames. 
This voting-based assignment ensures a one-to-one mapping 
between superpoints and instances, producing a unified 
and consistent 3D instance segmentation.

\section{Experiments} \label{experiments}
 
\subsection{Experimental Setup}
\textbf{Datasets.} We evaluated the robustness of our method on two benchmark datasets: ScanNet200~\cite{rozenberszki2022language} and ScanNet++~\cite{yeshwanth2023scannet++}. These datasets offer a diverse range of environments for testing 3D instance segmentation. For our experiments, we used the validation sets of ScanNet200 and ScanNet++.

\textbf{Metrics.} We assess the class-agnostic instance segmentation quality using the average precision (AP) metric, which is standard for instance segmentation evaluation. This metric disregards semantic labels and focuses solely on mask quality. AP is computed at different mask overlap thresholds—50\% and 25\%—and averaged across IoU thresholds from 0.5 to 0.95 in 0.05 increments. This provides a comprehensive assessment of segmentation accuracy across varying overlap conditions.

\textbf{Implementation Details.}  
Our method processes posed RGB-D frames from each scene sequence. We experiment with both SAM~\cite{kirillov2023segment} and CropFormer~\cite{qi2022high} as the 2D mask generator. 
For temporal tracking, we set the frame queue length to \(q_{\text{max}} = 5\) and use a 2D IoU threshold of \(\tau_{\text{IoU2D}} = 0.8\) for mask matching. 
3D superpoints are computed once for the entire scene using the method of 
Felzenszwalb and Huttenlocher~\cite{felzenszwalb2004efficient} 
and projected into each frame for association with 2D instance masks. 
Merging of instances across frames is performed based on a 3D IoU threshold 
of \(\tau_{\text{IoU3D}} = 0.6\), followed by refinement using a 3D IoMin threshold 
of \(\tau_{\text{IoMin3D}} = 0.8\) to resolve overlapping assignments. 
Temporal co-occurrence thresholding is set to \(\tau_{\text{co}} = 0.5\) 
to distinguish overlapping but distinct objects from 
partially duplicated observations.

\begin{table}[t!]
\vspace{+4mm}
\centering
\caption{Class-agnostic instance segmentation on ScanNet200~\cite{rozenberszki2022language}.}
\label{tab:scannet200}
\begin{threeparttable}
\begin{tabular}{@{}lccccc@{}}
\toprule
\textbf{Model} & \textbf{2D Model} & \textbf{AP} & \textbf{AP$_{50}$} & \textbf{AP$_{25}$} \\ \midrule
\textcolor{gray}{Mask3D†~\cite{schult2023mask3d}} & \textcolor{gray}{-} & \textcolor{gray}{39.7} & \textcolor{gray}{53.6} & \textcolor{gray}{62.5}\\
\midrule
SAMPro3D~\cite{xu2023sampro3d} & SAM~\cite{kirillov2023segment} & 18.0 & 32.8 & 56.1 \\
SAM3D~\cite{yang2023sam3d} & SAM~\cite{kirillov2023segment} & 20.2 & 35.7 & 55.5 \\
Open3DIS~\cite{nguyen2024open3dis} & 
SAM~\cite{kirillov2023segment} & 31.5 & 45.3 & 51.1 \\
MaskClustering~\cite{yan2024maskclustering}      & CropFormer~\cite{qi2022high} & 19.2 & 36.6 & 51.7 \\
OV-MAP~\cite{kim2024ov} & CropFormer~\cite{qi2022high} & 29.9 & 49.4 & 67.5\\
\midrule
\multirow{2}{*}{\textbf{Ours}} & SAM~\cite{kirillov2023segment} & 30.1 & 44.9 & 57.8 \\
 & CropFormer~\cite{qi2022high} & \textbf{33.2} & \textbf{52.1} & \textbf{69.2} \\
\bottomrule
\end{tabular}
\vspace{0.5mm}
\begin{tablenotes}
\footnotesize
\item[†]: Fully supervised model with 3D ground truth masks
\end{tablenotes}
\end{threeparttable}
\vspace{-5mm}
\end{table}

\subsection{Quantitative Results}
\begin{figure*} [ht!]
\vspace{+2mm}
\begin{center}
\includegraphics[width=0.95\textwidth]{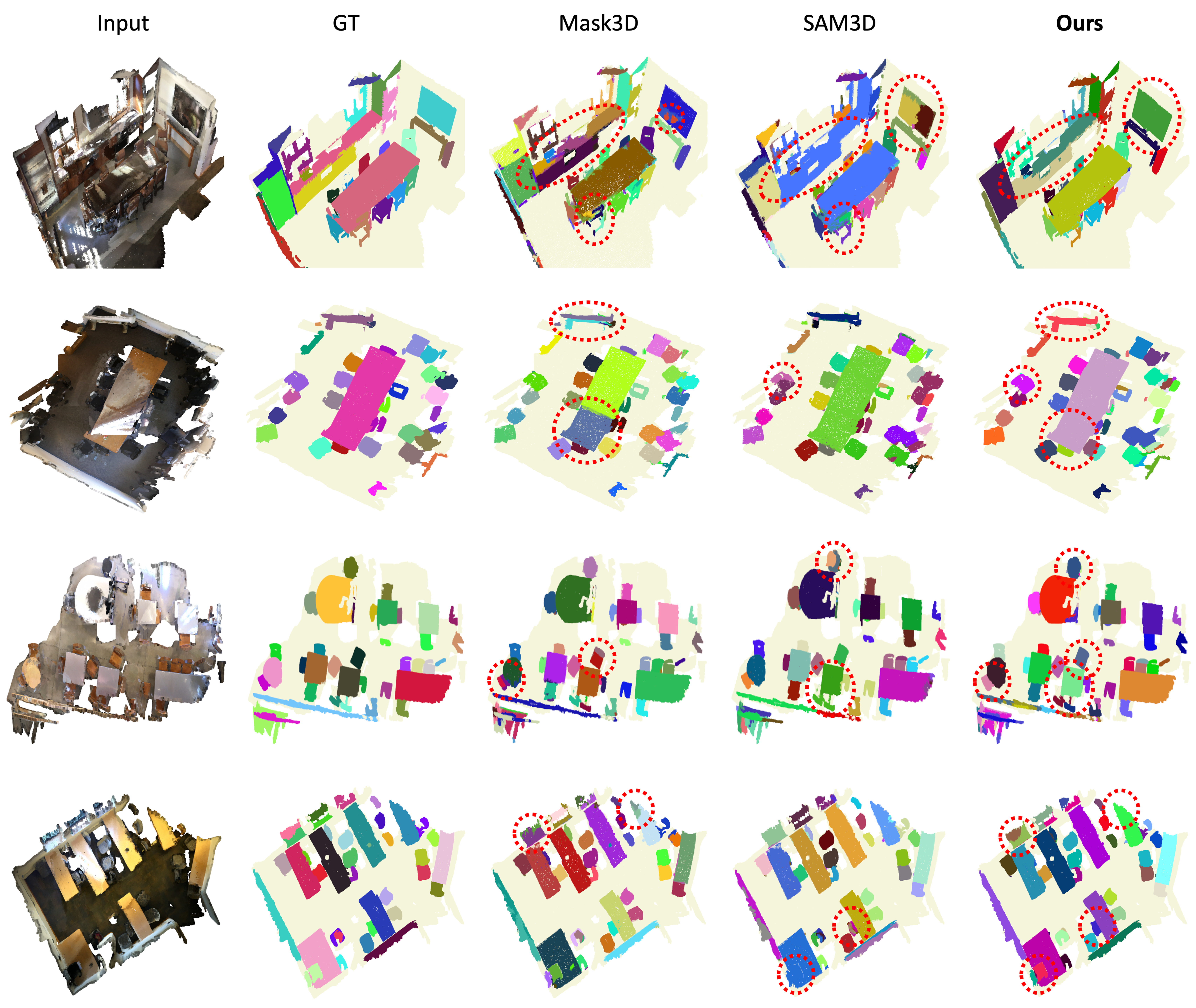}
\vspace{-2mm}
\caption{\textbf{Comparison of Class-Agnostic 3D Instance Segmentation on ScanNet200.}  
Visualization of 3D instance segmentation results. From left to right: input, ground truth (GT), Mask3D~\cite{schult2023mask3d}, SAM3D~\cite{yang2023sam3d}, and our method (CDIS). CDIS demonstrates improved segmentation accuracy, particularly in distinguishing object boundaries and reducing over-segmentation, as highlighted with red circles.}
\label{fig_result_ex}
\vspace{-6mm}
\end{center}
\end{figure*}

\textbf{ScanNet200 Evaluation.} Table~\ref{tab:scannet200} presents the performance of our method on the ScanNet200 benchmark. Our CropFormer-based model achieves an AP of 33.2, surpassing OV-MAP~\cite{kim2024ov} and Open3DIS~\cite{nguyen2024open3dis}, while also reporting the highest AP$_{50}$ (52.1) and AP$_{25}$ (69.2) among 2D-3D approaches. Compared to its SAM-based counterpart, our CropFormer-based model yields more coherent instance groupings in 3D, attributed to its stronger entity-level segmentation capability. Both versions benefit from CDIS’s cross-dimensional reasoning, where temporally tracked 2D masks are used to associate and consolidate 3D superpoints, resulting in stable and consistent instance labels across frames. While Mask3D~\cite{schult2023mask3d}, a fully supervised model, achieves the highest AP (39.7), it relies on dense 3D annotations. In contrast, our model achieves 33.2 AP without any 3D supervision, highlighting its practicality in annotation-scarce settings and its generalization to novel object categories.

\textbf{ScanNet++ Evaluation.} On the more challenging ScanNet++ dataset (Table~\ref{tab:scannet++}), our CropFormer-based model achieves an AP of 28.2 and AP$_{50}$ of 43.7, demonstrating strong performance comparable to state-of-the-art methods such as MaskClustering~\cite{yan2024maskclustering} and Open3DIS~\cite{nguyen2024open3dis}. In AP$_{25}$, our model scores 54.3, which is on par with MaskClustering~\cite{yan2024maskclustering} (54.7), showing that both approaches perform similarly at this level. These results validate the effectiveness of our cross-dimensional segmentation framework in handling complex 3D environments, particularly in improving instance-level consistency. Moreover, the consistent performance across datasets demonstrates CDIS’s robustness to scene variation, making it well-suited for robotic tasks in unfamiliar environments.

\subsection{Qualitative Results}
As shown in Fig.~\ref{fig_result_ex}, we present qualitative examples of our class-agnostic 3D instance segmentation on ScanNet200. By leveraging 2D instance segmentation models, CDIS delivers high-quality 3D instance segmentation without requiring 3D-annotated training data. Mask3D~\cite{schult2023mask3d}, trained on ScanNet200 with full 3D supervision, achieves strong results but struggle with fragmented instances and occluded objects. Despite not being trained on 3D-annotated data, CDIS performs comparably or even outperforms Mask3D in challenging scenarios, particularly in preserving object boundaries and handling occlusions. Additionally, unlike SAM3D~\cite{yang2023sam3d}, which merges projected 2D masks in 3D space, CDIS tracks and refines 2D masks, which are then used to label 3D superpoints through projection overlap, reducing segmentation errors such as over-segmentation and object merging. As highlighted with red circles in Fig.~\ref{fig_result_ex}, CDIS produces results that closely match the ground truth, demonstrating its robustness in complex indoor scenes.

\begin{table}[t!]
\centering
\caption{Class-agnostic instance segmentation on ScanNet++~\cite{yeshwanth2023scannet++}.}
\label{tab:scannet++}
\begin{threeparttable}
\begin{tabular}{@{}lccccc@{}}
\toprule
\textbf{Model} & \textbf{2D Model} & \textbf{AP} & \textbf{AP$_{50}$} & \textbf{AP$_{25}$} \\ \midrule
\textcolor{gray}{Mask3D†~\cite{schult2023mask3d}} & \textcolor{gray}{-} & \textcolor{gray}{22.8} & \textcolor{gray}{33.3} & \textcolor{gray}{45.7}\\
\midrule
SAM3D~\cite{yang2023sam3d} & SAM~\cite{kirillov2023segment} & 7.2 & 14.2 & 29.4 \\
Segment3D~\cite{huang2023segment3d} & SAM~\cite{kirillov2023segment} & 19.0 & 29.7 & 41.6\\
Open3DIS~\cite{nguyen2024open3dis} & SAM~\cite{kirillov2023segment} & 20.7 & 38.6 & 47.1\\
MaskClustering~\cite{yan2024maskclustering} & CropFormer~\cite{qi2022high} & 27.9 & 42.8 & \textbf{54.7}\\
\midrule
\multirow{2}{*}{\textbf{Ours}} & SAM~\cite{kirillov2023segment} & 22.9 & 35.4 & 46.7 \\
 & CropFormer~\cite{qi2022high} & \textbf{28.2} & \textbf{43.7} & 54.3 \\
\bottomrule
\end{tabular}
\vspace{0.5mm}
\begin{tablenotes}
\footnotesize
\item[†]: Fully supervised model with 3D ground truth masks
\end{tablenotes}
\end{threeparttable}
\vspace{-3mm}
\end{table}

\subsection{Real-World Experiments}
To validate the practical applicability of CDIS, we performed real-world experiments 
using RGB-D data captured in indoor environments (home and office settings). 
Camera poses and synchronized RGB-D frames were obtained using the 
RTAB-Map visual SLAM method~\cite{labbe2019rtab}, 
and the fused scene geometry was reconstructed using a 
Truncated Signed Distance Function volume integration~\cite{newcombe2011kinectfusion}. 
The captured environments contained various objects, 
such as chairs, tables, shelves with small items, and reflective surfaces 
under different lighting conditions.

\begin{figure} [t!]
\vspace{+3mm}
\begin{center}
\includegraphics[width=\columnwidth]{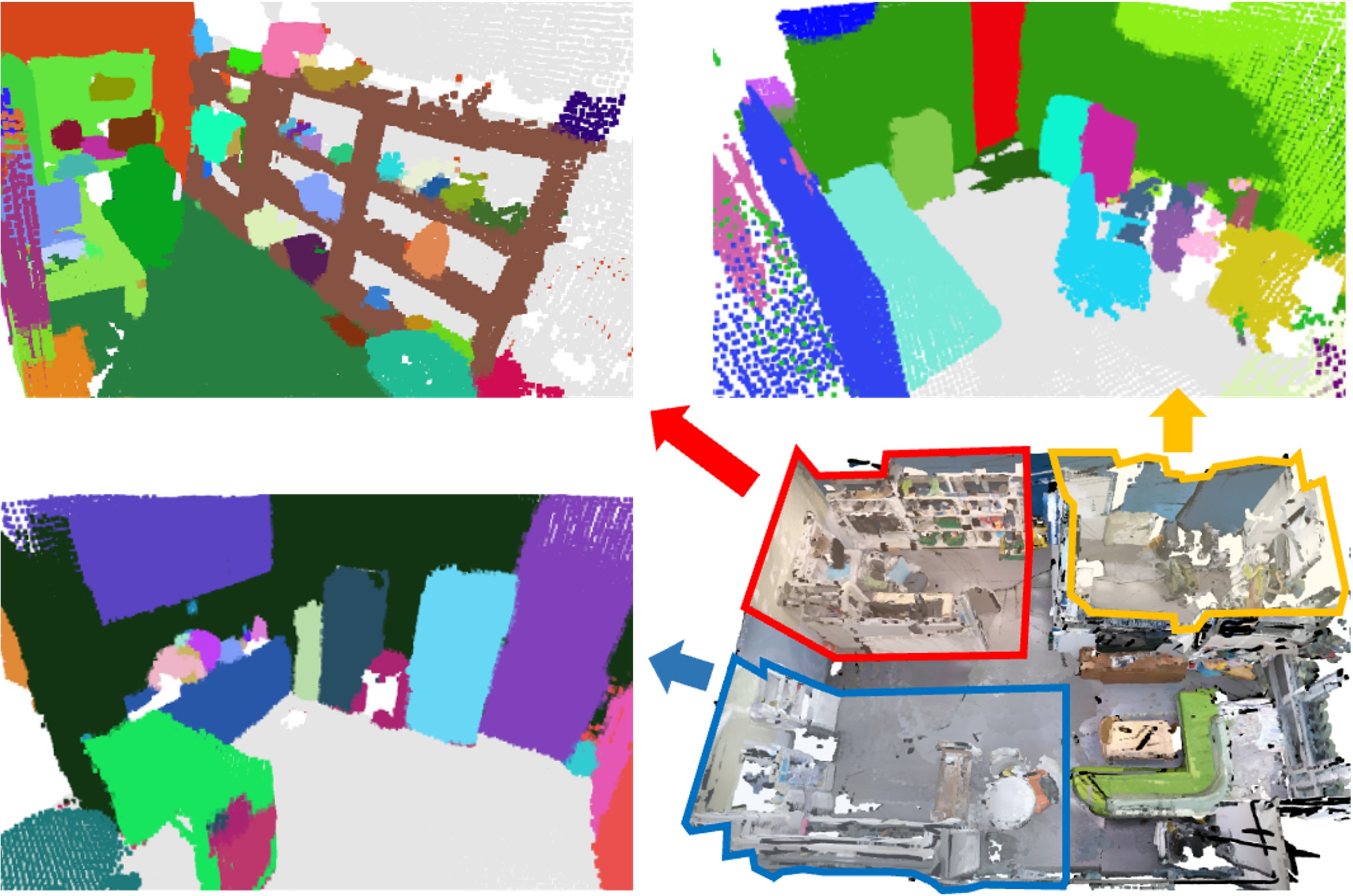}
\vspace{-3mm}
\caption{\textbf{Example of Real-World Class-Agnostic 3D Instance Segmentation.} CDIS performance on real-world data, demonstrating class-agnostic instance segmentation across large and diverse indoor environments, including an office (red), a bedroom (yellow), and a kitchen (blue). Our method effectively segments cluttered and complex scenes without 3D training.}
\label{real_ex}
\vspace{-6mm}
\end{center}
\end{figure}

CDIS successfully segmented most objects, particularly small items on shelves, which are essential for robotic grasping and manipulation. However, segmentation accuracy decreased near windows with strong reflections due to inconsistent 2D mask predictions. Fig.~\ref{real_ex} illustrates successful segmentations in challenging areas. These experiments demonstrate CDIS's potential for integration into robotic perception systems for object manipulation, autonomous navigation, and mapping, enabling robust scene understanding without requiring 3D-specific training.

\bibliographystyle{ieeetr}
\bibliography{ref}

@inproceedings{kirillov2023segment,
  title={Segment anything},
  author={Kirillov, Alexander and Mintun, Eric and Ravi, Nikhila and Mao, Hanzi and Rolland, Chloe and Gustafson, Laura and Xiao, Tete and Whitehead, Spencer and Berg, Alexander C and Lo, Wan-Yen and others},
  booktitle={Proceedings of the IEEE/CVF International Conference on Computer Vision},
  pages={4015--4026},
  year={2023}
}

@article{qi2022high,
  title={High-quality entity segmentation},
  author={Qi, Lu and Kuen, Jason and Guo, Weidong and Shen, Tiancheng and Gu, Jiuxiang and Jia, Jiaya and Lin, Zhe and Yang, Ming-Hsuan},
  journal={arXiv preprint arXiv:2211.05776},
  year={2022}
}

@inproceedings{cheng2022masked,
  title={Masked-attention mask transformer for universal image segmentation},
  author={Cheng, Bowen and Misra, Ishan and Schwing, Alexander G and Kirillov, Alexander and Girdhar, Rohit},
  booktitle={Proceedings of the IEEE/CVF conference on computer vision and pattern recognition},
  pages={1290--1299},
  year={2022}
}

@article{cheng2021per,
  title={Per-pixel classification is not all you need for semantic segmentation},
  author={Cheng, Bowen and Schwing, Alex and Kirillov, Alexander},
  journal={Advances in neural information processing systems},
  volume={34},
  pages={17864--17875},
  year={2021}
}

@inproceedings{rozenberszki2022language,
    title={Language-Grounded Indoor 3D Semantic Segmentation in the Wild},
    author={Rozenberszki, David and Litany, Or and Dai, Angela},
    booktitle = {Proceedings of the European Conference on Computer Vision ({ECCV})},
    year={2022}
}

@inproceedings{schult2023mask3d,
  title={Mask3d: Mask transformer for 3d semantic instance segmentation},
  author={Schult, Jonas and Engelmann, Francis and Hermans, Alexander and Litany, Or and Tang, Siyu and Leibe, Bastian},
  booktitle={2023 IEEE International Conference on Robotics and Automation (ICRA)},
  pages={8216--8223},
  year={2023},
  organization={IEEE}
}

@article{takmaz2023openmask3d,
  title={Openmask3d: Open-vocabulary 3d instance segmentation},
  author={Takmaz, Ay{\c{c}}a and Fedele, Elisabetta and Sumner, Robert W and Pollefeys, Marc and Tombari, Federico and Engelmann, Francis},
  journal={arXiv preprint arXiv:2306.13631},
  year={2023}
}

@article{yang2023sam3d,
  title={Sam3d: Segment anything in 3d scenes},
  author={Yang, Yunhan and Wu, Xiaoyang and He, Tong and Zhao, Hengshuang and Liu, Xihui},
  journal={arXiv preprint arXiv:2306.03908},
  year={2023}
}

@inproceedings{yan2024maskclustering,
  title={Maskclustering: View consensus based mask graph clustering for open-vocabulary 3d instance segmentation},
  author={Yan, Mi and Zhang, Jiazhao and Zhu, Yan and Wang, He},
  booktitle={Proceedings of the IEEE/CVF Conference on Computer Vision and Pattern Recognition},
  pages={28274--28284},
  year={2024}
}

@article{straub2019replica,
  title={The Replica dataset: A digital replica of indoor spaces},
  author={Straub, Julian and Whelan, Thomas and Ma, Lingni and Chen, Yufan and Wijmans, Erik and Green, Simon and Engel, Jakob J and Mur-Artal, Raul and Ren, Carl and Verma, Shobhit and others},
  journal={arXiv preprint arXiv:1906.05797},
  year={2019}
}

@article{chang2017matterport3d,
  title={Matterport3d: Learning from rgb-d data in indoor environments},
  author={Chang, Angel and Dai, Angela and Funkhouser, Thomas and Halber, Maciej and Niessner, Matthias and Savva, Manolis and Song, Shuran and Zeng, Andy and Zhang, Yinda},
  journal={arXiv preprint arXiv:1709.06158},
  year={2017}
}

@inproceedings{hou20193d,
  title={3d-sis: 3d semantic instance segmentation of rgb-d scans},
  author={Hou, Ji and Dai, Angela and Nie{\ss}ner, Matthias},
  booktitle={Proceedings of the IEEE/CVF conference on computer vision and pattern recognition},
  pages={4421--4430},
  year={2019}
}

@inproceedings{yin2024sai3d,
  title={Sai3d: Segment any instance in 3d scenes},
  author={Yin, Yingda and Liu, Yuzheng and Xiao, Yang and Cohen-Or, Daniel and Huang, Jingwei and Chen, Baoquan},
  booktitle={Proceedings of the IEEE/CVF Conference on Computer Vision and Pattern Recognition},
  pages={3292--3302},
  year={2024}
}

@inproceedings{rozenberszki2024unscene3d,
  title={Unscene3d: Unsupervised 3d instance segmentation for indoor scenes},
  author={Rozenberszki, David and Litany, Or and Dai, Angela},
  booktitle={Proceedings of the IEEE/CVF Conference on Computer Vision and Pattern Recognition},
  pages={19957--19967},
  year={2024}
}

@article{xu2023sampro3d,
  title={Sampro3d: Locating sam prompts in 3d for zero-shot scene segmentation},
  author={Xu, Mutian and Yin, Xingyilang and Qiu, Lingteng and Liu, Yang and Tong, Xin and Han, Xiaoguang},
  journal={arXiv preprint arXiv:2311.17707},
  year={2023}
}

@inproceedings{yeshwanth2023scannet++,
  title={Scannet++: A high-fidelity dataset of 3d indoor scenes},
  author={Yeshwanth, Chandan and Liu, Yueh-Cheng and Nie{\ss}ner, Matthias and Dai, Angela},
  booktitle={Proceedings of the IEEE/CVF International Conference on Computer Vision},
  pages={12--22},
  year={2023}
}

@inproceedings{kim2024ov,
  title={OV-MAP: Open-Vocabulary Zero-Shot 3D Instance Segmentation Map for Robots},
  author={Kim, Juno and Park, Yesol and Yoon, Hye-Jung and Zhang, Byoung-Tak},
  booktitle={2024 IEEE/RSJ International Conference on Intelligent Robots and Systems (IROS)},
  pages={13780--13786},
  year={2024},
  organization={IEEE}
}

@inproceedings{lu2023ovir,
  title={Ovir-3d: Open-vocabulary 3d instance retrieval without training on 3d data},
  author={Lu, Shiyang and Chang, Haonan and Jing, Eric Pu and Boularias, Abdeslam and Bekris, Kostas},
  booktitle={Conference on Robot Learning},
  pages={1610--1620},
  year={2023},
  organization={PMLR}
}

@inproceedings{nguyen2024open3dis,
  title={Open3dis: Open-vocabulary 3d instance segmentation with 2d mask guidance},
  author={Nguyen, Phuc and Ngo, Tuan Duc and Kalogerakis, Evangelos and Gan, Chuang and Tran, Anh and Pham, Cuong and Nguyen, Khoi},
  booktitle={Proceedings of the IEEE/CVF Conference on Computer Vision and Pattern Recognition},
  pages={4018--4028},
  year={2024}
}

@article{guo2023sam,
  title={Sam-guided graph cut for 3d instance segmentation},
  author={Guo, Haoyu and Zhu, He and Peng, Sida and Wang, Yuang and Shen, Yujun and Hu, Ruizhen and Zhou, Xiaowei},
  journal={arXiv preprint arXiv:2312.08372},
  year={2023}
}

@article{huang2023segment3d,
  title={Segment3d: Learning fine-grained class-agnostic 3d segmentation without manual labels},
  author={Huang, Rui and Peng, Songyou and Takmaz, Ayca and Tombari, Federico and Pollefeys, Marc and Song, Shiji and Huang, Gao and Engelmann, Francis},
  journal={arXiv preprint arXiv:2312.17232},
  year={2023}
}

@article{felzenszwalb2004efficient,
  title={Efficient graph-based image segmentation},
  author={Felzenszwalb, Pedro F and Huttenlocher, Daniel P},
  journal={International journal of computer vision},
  volume={59},
  number={2},
  pages={167--181},
  year={2004},
  publisher={Springer}
}

@inproceedings{ngo2023isbnet,
  title={Isbnet: a 3d point cloud instance segmentation network with instance-aware sampling and box-aware dynamic convolution},
  author={Ngo, Tuan Duc and Hua, Binh-Son and Nguyen, Khoi},
  booktitle={Proceedings of the IEEE/CVF Conference on Computer Vision and Pattern Recognition},
  pages={13550--13559},
  year={2023}
}

@article{labbe2019rtab,
  title={RTAB-Map as an open-source lidar and visual simultaneous localization and mapping library for large-scale and long-term online operation},
  author={Labb{\'e}, Mathieu and Michaud, Fran{\c{c}}ois},
  journal={Journal of field robotics},
  volume={36},
  number={2},
  pages={416--446},
  year={2019},
  publisher={Wiley Online Library}
}

@inproceedings{newcombe2011kinectfusion,
  title={Kinectfusion: Real-time dense surface mapping and tracking},
  author={Newcombe, Richard A and Izadi, Shahram and Hilliges, Otmar and Molyneaux, David and Kim, David and Davison, Andrew J and Kohi, Pushmeet and Shotton, Jamie and Hodges, Steve and Fitzgibbon, Andrew},
  booktitle={2011 10th IEEE international symposium on mixed and augmented reality},
  pages={127--136},
  year={2011},
  organization={Ieee}
}

@inproceedings{kolodiazhnyi2024oneformer3d,
  title={Oneformer3d: One transformer for unified point cloud segmentation},
  author={Kolodiazhnyi, Maxim and Vorontsova, Anna and Konushin, Anton and Rukhovich, Danila},
  booktitle={Proceedings of the IEEE/CVF Conference on Computer Vision and Pattern Recognition},
  pages={20943--20953},
  year={2024}
}

@inproceedings{park2025dafusion,
  title={{DA-Fusion}: Deformable Attention-based RGB-D Fusion Transformer for Unseen Object Instance Segmentation},
  author={Park, Yesol and Yoon, Hye-Jung and Kim, Juno and Zhang, Byoung-Tak},
  booktitle={2025 IEEE International Conference on Robotics and Automation (ICRA)},
  pages={7490--7496},
  year={2025},
  organization={IEEE}
}

@inproceedings{huang2024openins3d,
  title={Openins3d: Snap and lookup for 3d open-vocabulary instance segmentation},
  author={Huang, Zhening and Wu, Xiaoyang and Chen, Xi and Zhao, Hengshuang and Zhu, Lei and Lasenby, Joan},
  booktitle={European Conference on Computer Vision},
  pages={169--185},
  year={2024},
  organization={Springer}
}
\end{document}